# Understanding the Skills Gap between Higher Education and Industry in the UK in Artificial Intelligence Sector

*Khushi Jaiswal[1], Ievgeniia Kuzminykh[1,2], Sanjay Modgil[1,3]*
*[1] King's College London, UK*
*[2] Kharkiv National University of Radio Electronics, Ukraine*
*[3] University of Milan, Italy*
*Correspondence: ievgeniia.kuzminykh@kcl.ac.uk, King's College London, BH(N)6.06 Bush House Strand campus, 30 Aldwych, London, WC2B 4BG*

## Abstract

As Artificial Intelligence (AI) changes how businesses work, there's a growing need for people who can work in this sector. This paper investigates how well universities in United Kingdom offering courses in AI, prepare students for jobs in the real world. To gain insight into the differences between university curricula and industry demands we review the contents of taught courses and job advertisement portals. By using custom data scraping tools to gather information from job advertisements and university curricula, and frequency and Naive Bayes classifier analysis, this study will show exactly what skills industry is looking for. In this study we identified 12 skill categories that were used for mapping. The study showed that the university curriculum in the AI domain is well balanced in most technical skills, including Programming and Machine learning subjects, but have a gap in Data Science and Maths and Statistics skill categories.

**Index words:** Artificial intelligence, job advert, curriculum development, information systems, machine learning, skill requirements, Clay

## Introduction

Today's fast-changing world of artificial intelligence (AI) technologies is rapidly transforming industry. As AI becomes more crucial in businesses, there is a growing need for skilled professionals. A LinkedIn News (2023) study on fastest-growing jobs in 2023, stated that positions similar to those held by Machine Learning Engineers were recognised as one of the most prominent (top 20) jobs emerging over the past 6 years, with an increase of over 74% in demand for roles driven by the expansion of big data and emergence of new technologies. In a similar report from 2024, LinkedIn News (2024) put the new role of Artificial Intelligence Engineers as one of top 10 careers in the UK. This demand has led to the introduction of AI courses in universities.

It is often difficult to find students with perfect resumes and skill sets for such jobs (Romeo, 2020). Even in the extensive domain of the UK space industry, research has revealed that a significant majority, 95% to be exact, of space organisations encounter skill-related challenges. Amongst these organisations, nearly a quarter express a distinct and specific requirement for AI skills, surpassing the demand for expertise in any other technical domain (Li et al., 2023; UK Space Agency, 2023).

A similar concern in respect of AI expertise shortage was voiced by the senior Vice President of human resources at Bosch, a leading tech company that uses and offers many AI applications, as stated in the article by Human Resource Executive (Romeo, 2020), and by Stewart Townsend, a highly skilled consultant with 20 years of experience in channel strategy, management, and growth through strategic partnerships (Townsend, 2024). Demand for AI experts is high, but there aren't enough skilled workers to meet it. This shortage affects many industries and the global economy. Why is there a shortage? There is no clear answer to this, although one might speculate that it is due to the rapidly evolving AI landscape, and universities and training programs aren't keeping up with these rapid changes. In this study we attempt to shed some light on the quality of education in AI. As observed by Townsend (2024), universities and training organisations need to improve the scope and quality of AI education, and focus on specialised AI training, in order to ensure that the shortage in AI skills will not worsen and hamper innovative deployment of AI in industries such as healthcare and finance.

The UK government has already taken steps to tackle the AI skills deficit. In Autumn 2020, the government introduced a degree conversion course program focused on AI and data science, providing support for students in the form of scholarships. This initiative's objective was to produce a minimum of 2,500 graduates within 3 years, with the intention of narrowing the skills gap (Department for Science, Innovation and Technology, 2021; Office for Students, 2023). According to a survey conducted by the Career Development Organisation (CRAC, 2024), which reviewed programmes in AI between April 2020 and March 2023, the initiative's funding has helped support 7,600 students enrolled on AI and data science conversion Masters courses. Reports suggest that the "programme has had a substantial positive effect on the number of postgraduate students" in AI in the UK, with extra opportunities being provided to women, black students and students with declared disabilities. Despite the success of the program, a segment head of the program has stated that looking forward to future academic years, universities should explore where additional work is needed to further decrease the present skills gap.

Following on from these recent developments, our study will aim to understand the skill and expertise gap between higher education and industry. Specifically, skills that Artificial Intelligence degrees equip students with, will be mapped to the skills required by UK-based jobs in the AI sector. To achieve the aim, we pose these three research questions:

> *RQ1: What skills are required by UK AI industry recruiters?*
>
> *RQ2: What skills are gained by students who have completed an AI degree in the UK?*
>
> *RQ3: Does university curricula offer a sufficient level of knowledge to match industry requirements in AI related jobs?*

To summarise, many studies (Han and Anderson, 2021; Department for Digital, Culture, Media and Sport, 2021; Sarin, 2019; Verma, Lamsal and Verma, 2022) have investigated the demands of industry, and analysed job adverts, but none have undertaken a mapping to programmes in higher education institutions (HEIs). Hence, provision of such a mapping, between skills provided by university courses and the skills that jobs in the AI sector require, is the key contribution of this study. Also, many studies were carried out for the USA, while few addressed the UK or Europe. Our report will focus explicitly on the UK. The methodology chosen for our study, combines successful elements from past works, with additional statistical evaluation of the collected data to see how relevant the findings are.

The rest of the paper is structured as follows. First, we will review existing investigations into industry demands for AI skills, and efforts to map these demands with qualifications offered by universities. Next, we present the methodology used in our study to collect and analyse the data from job adverts' portals and university curricula. Next, we present results and evaluate the findings along with any conclusions drawn from our analysis. Finally, we outline the conclusions and directions for future research.

## Research Background

Numerous research papers with comparable objectives have been published in the same domain as this study. A thorough examination of these works has significantly contributed to the development of our research methodology.

Verma, Lamsal and Verma (2022) discuss the skills requirements in AI and ML in job advertisements for the US market. Their work indicates that despite high demand, there is a shortage of skilled talent. The methodology employed in their report involves a rigorous process designed to scrape, map and analyse the skills requirements for ML and AI. The source for the job advertisements used by the report was the job platform indeed.com. The report employed both manual approaches and automated approaches to collecting the relevant job adverts. Python and its libraries were utilised for web scraping, enabling the extraction of job titles and descriptions. Content analysis and n-gram techniques were applied to extract relevant information from job descriptions. The n-grams enabled consideration of skills identified with not just single words but described by phrases of two or more words. A similar phrase analysis will also be conducted in our research.

A unique and rather ambitious aspect of the study by Verma, Lamsal and Verma, was the use of a framework for classifying skills into distinct categories. The framework consists of AI and ML categories, and skills are mapped onto one of the available categories. Additionally, the classification framework was validated by three independent evaluators to ensure the reliability of the skill categorisation. The analysis revealed a substantial demand for AI and ML professionals, particularly in California, Washington, and New York with figures showing that 34% of all job posts in AI and ML originating from California, 10% of all job posts in AI from Washington and 12% of all jobs in ML originating from New York.

Alternatives to this proposed classification framework can be found in other works (Pranckevicius and Marcinkevicius, 2017; Manning, Raghavan and Schütze, 2008) which is apart of classification performing as well a text analysis. The papers take into account the length and complexity of the words as well as the size of the data set used.

The study performed by Attwood and Williams (2023) investigated the skills gap in the cybersecurity industry. Their work outlines the significance of addressing the skills gap, especially in cyber-enabled disciplines like software engineering. Attwood and Williams aim to evaluate the potential of mapping job descriptions to the Cyber Security Body of Knowledge (CyBOK), using TF-IDF (Term Frequency-Inverse Document Frequency) representations. It is a numerical statistic that reflects the importance of a word in a document relative to a collection of document representations.

The researchers gathered information from a source that provides job listings in UK related cybersecurity and software engineering domains. The data includes details like job titles, descriptions, salary ranges, and locations. They categorised the job listings into different types based on their titles and descriptions. The main part of the study involved connecting the job descriptions to specific areas of knowledge outlined in the CyBOK. They used TF-IDF, which basically measures how important specific words are in a document, compared to a larger collection of documents. The final findings identify the top performing knowledge areas for the different job descriptions collected. For instance, a developer's job description usually excels in the Secure Software Lifecycle CyBOK knowledge area.

Common to all these related works is the use of web scraping to collect data. The success of web scraping is further emphasised in Han and Anderson (2021), which conducted an in-depth comparative analysis of different data collection methods for online market research in the hospitality domain. The paper rightly notes that the internet provides valuable opportunities for collecting and studying consumer choices. The report then goes on to compare web scraping (Glez-Peña et al., 2014; Khder, 2021), programming interfaces (APIs), taking surveys and simulating online behaviours in lab settings. Table 1 provides a comparison of all the stated methods of data collection and highlights how API access can be challenging, limited, or costly, and simulation studies may lack realism. While outsourcing web scraping to third parties is an option, there are potential pitfalls like the lack of care when collecting data. To conclude, web scraping is cost-effective and provides large-scale, real-time data, in contrast to the challenges posed by conducting surveys.

**Table 1: Comparison of Common Data Collection Methods (Han and Anderson, 2021).**

|  | **Scraped Data** | **Commercial Web Scraping Service** | **API** | **Survey** |
|---|---|---|---|---|
| Cost | Low | Medium | Low/Medium | High |
| Sample frame | Website users | Website users | Website users | Flexible |
| Customizability of variables | Medium | Low | Low | High |
| Ease of frequent collection | Easy | Moderate | Easy | Hard |
| Data type | Behavioural | Behavioural | Behavioural | Attitudinal |
| Limitations | Time and programming skills | Data may not be suitable to the researcher's need in terms of variables or content | Limited availability | Time and programming skills |

The study conducted by Sarin (2019) identifies a disparity between the skills perceived as important by students and those valued by employers. Because the skills gap here is being

*Understanding the Skills Gap between Higher Education and Industry in the UK in AI Sector*identified in terms of peoples' opinions, the methodology used differs to that used by other studies. Sarin's main study involved collecting primary data from 230 graduate and post-graduate students in the Delhi region. Convenience sampling was used, which means participants were selected based on their easy availability. In the second part of the study, primary data was collected from 50 human resource (HR) executives in the Delhi region. These HR executives were responsible for recruitment from higher education institutes. Both students and HR executives were asked to rate the importance of skills on a scale ranging from 1 (least important) to 5 (most important). This allowed for quantifying perceptions of skill importance. The results of the study showed that students often spend a lot of their time on quantitative topics while employers believe that communication skills and domain knowledge is also very important.

In addition, Department for Digital, Culture, Media and Sport report (2021) from the official government website looks extensively at the UK AI labour market, and also emphasises a skill shortage. It discusses difficulties in filling vacancies and identifies where the majority of jobs adverts are posted. The study employed both statistical analysis of the AI jobs posted in the period between January and December 2020, and elicitation methods through interviewing the companies and organisations somehow connected to the AI sector. The companies pointed to a skills gap in technical skills such as AI concepts and algorithms, programming languages, software and systems engineering, as well as in soft skills and privacy and ethics. The importance of soft skills is further illustrated in Gardiner et al. (2018), in which the authors conclude that soft skills remain highly valued, in addition to emerging hard technological skills.

In conclusion, the research reviewed above have provided background and insights for this paper's research. We use the concept of grouping skills into n-grams presented in (Verma, Lamsal and Verma, 2022) before conducting a frequency analysis and a mapping analysis similar to that used by (Attwood and Williams, 2023) for university programmes in cybersecurity. Many studies have extracted and analysed data from job platforms, but only (Attwood and Williams, 2023) performed the actual mapping of skills from university degrees to the skills that are required by job recruiters. However, the study investigated the cybersecurity sector and had predefined categories for mapping, whereas in our study we first define the categories of skills in AI that could be used in analysis; a unique feature of our work. This mapping will hopefully uncover further disparities, ultimately helping us to identify a skills gap.

## Research Methodology

Figure 1 describes the methodology used for our study. First, the skills that employers required for AI related jobs were scraped and stored. This data further was cleaned, and the frequency of popular skills recorded. The second dataset, collating the skills that are acquired by students who have completed AI related degrees at university, was created in the same way. We then performed a mapping between the two datasets to see if any pattern or correlation can be inferred and presented the findings.

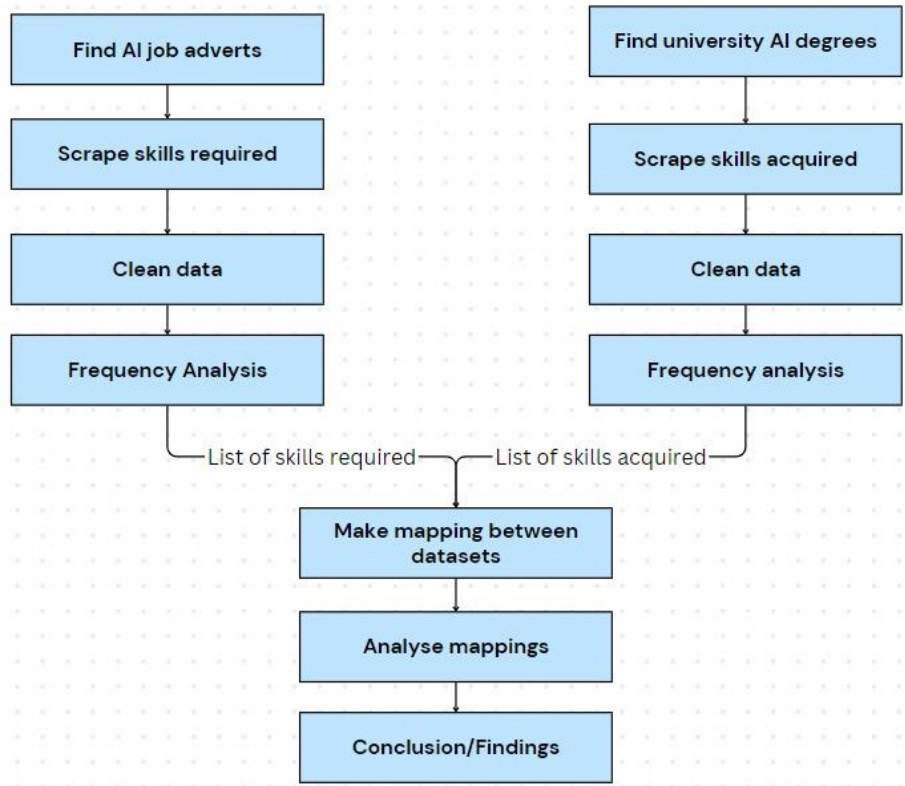

**Figure 1: Summary of research methodology used.**

## Data Collection

We used LinkedIn as a job advertisement platform because it gives a lot of detailed information about the companies posting job adverts. Each job description tends to be relatively thorough, giving a good sense of the company's profile. Additionally, LinkedIn lists exactly what they are looking for in a candidate, i.e. the required skills. Another important reason for choosing LinkedIn is that it enables use of filters to help narrow down searches, which makes finding the right job postings a lot easier and more efficient.

We used a chrome extension called Clay to collect data from LinkedIn. This extension is widely used for automation of recruiting process and provides the customization options for the scraping process, such as specifying the data fields to extract and setting up filters. Automated scrapping is the most commonly used technique seen in previous studies (Attwood and Williams, 2023; Han and Anderson, 2021; Sarin, 2019; Verma, Lamsal and Verma, 2022). The process of data scrapping is shown in Figure 2. LinkedIn data scraping is legal, but it is still not encouraged by the platform (Chen, 2024). In addition, LinkedIn uses algorithms to detect unauthorised scraping by monitoring actions that do not seem "human."

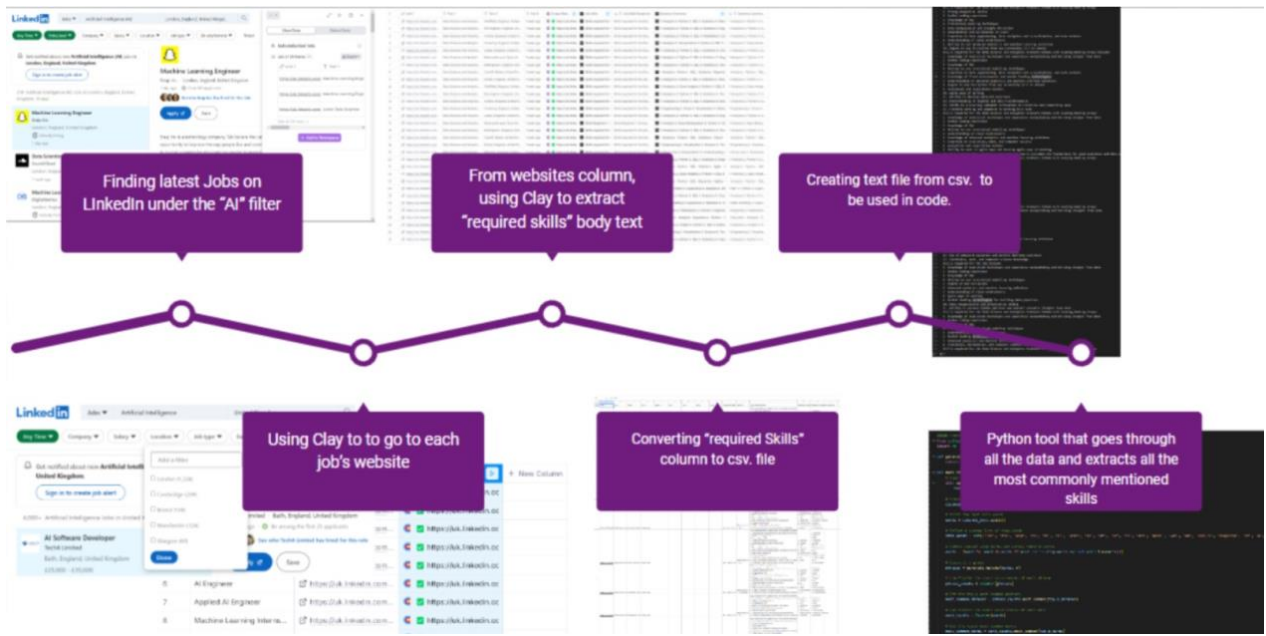

**Figure 2: The collection of data from LinkedIn using the extension Clay.**

On the other hand, online universities websites are a highly reliable source for the skills that students gain while at that institution. These sites are very structured, easy to navigate and frequently updated with the latest module changes. Moreover, each university usually only has a single website with all the information in one place. This makes data collection easier as we do not have to compare among multiple other sources. The only major decision to make at this stage was which universities to select for the data collection. According to the latest officially documented numbers from the UK government published in Policy paper under the title "AI Sector Deal" (Government of the UK, 2019), there were only 26 out of 166 universities (15.6%) in the UK in 2017 offering undergraduate courses in AI. Since the government actions described in the introduction have been applied the number increased to 61 UK universities offering undergraduate degree in AI, as of March 2024 (see Figure 3).

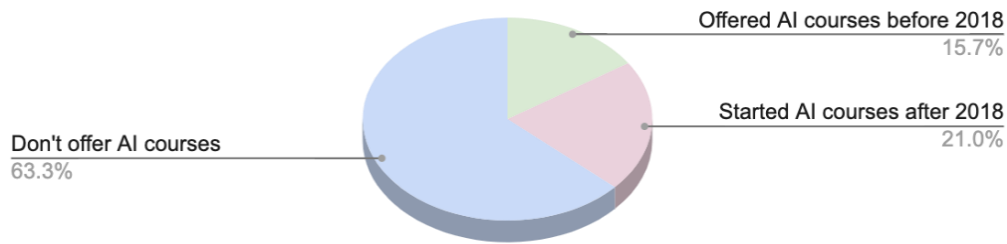

**Figure 3: Proportion of universities that offer undergraduate AI courses.**

Our study will focus on universities that do offer AI related courses; in particular highly ranked UK universities according to university QS ranking (QS Top Universities, 2023). The rationale behind this decision is that if our analysis identifies a skills gap in respect of AI courses offered by the top UK universities, then this is at least indicative of the highest quality AI courses offered in the UK, and so suggests at least a lower bound on the size of the skills gap (one would anticipate a higher skills gap if taking into account AI courses offered by university degrees that are not as highly ranked).

After scrapping of web sites was completed, we obtained two datasets: a job dataset with skills from 158 AI related job adverts and a university dataset with skills from 30 different university AI courses.

### *Data Processing*

Since both datasets are unstructured, we performed a cleansing of data associated with removing stop words, generating n-grams (n-grams are sequences of n consecutive words) and counting the number of occurrences for each skill. We used the Python Natural Language Toolkit (NLTK) to clean the text by removing all symbols, numbers and whitespaces. The remaining strings are then converted to lowercase, and then split into individual words. Then, we extracted word phrases of length from 1 to 5 to identify the most common words that represented the skills.

### *Data Analysis*

The word phrase frequency analysis was conducted on both datasets, collecting the top 400 most commonly occurring phrases for each of the 5 different lengths of phrases, giving us a total of 1000 skills in each dataset. Next, we provide the mapping between these two datasets. The mapping will allow us to find any patterns or discrepancies between the two sets indicating a skills gap or the absence of it.

There are many ways of achieving a good mapping. This study used the method of categorisation. For classification of extracted skills into categories we used the Naive Bayes model, which is a probabilistic learning model normally used for classification tasks, including text classification (Verma, Lamsal and Verma, 2022). In our study we present 12 distinct categories for skills classification, shown in Table 2. Each has some of the most common related skills identified from the job descriptions.

To evaluate the proposed model, we split the obtained dataset into two parts: 90% of the training data is used to train the model and the remaining 10% is used as the testing set. The classification report showed how well the trained model performs for each of the 12 predefined categories. Our model showed an average precision of 87% and an average F1-score of 82%, indicating that it can be considered as highly reliable.

**Table 2: The skill categories, their examples and abbreviations**

| Category | Common Skills Found | Abbreviation |
|---|---|---|
| Programming & software development | Languages<br>Algorithms<br>Version control | PROG |
| Machine learning | Supervised & unsupervised learning<br>Neural networks<br>Deep learning | ML |
| Data science and analytics | Data cleaning & processing<br>Predictive modelling<br>Visualisation | DATA |
| Engineering | Agile<br>System design<br>Distributed computing | ENG |
| Business and management | Project management<br>Cost-benefit analysis | BIZ |
| Cloud and technologies | AWS<br>Docker<br>Big data | CLOUD |
| Soft skills (communication & collaboration) | Communication<br>Collaboration<br>Problem-solving | SOFT |
| Maths and statistics | Probability<br>Linear algebra<br>Optimization | MATHS |
| Research and development | Literature review<br>Prototyping<br>Innovation | R&D |
| Industry specific knowledge | Domain expertise<br>Industry challenges<br>Relevant datasets | IND |
| Tools and technologies | Jupiter notebook<br>TensorFlow<br>Pandas | TOOLS |
| Ethics | Fairness<br>Transparency<br>Bias detection | ETH |

# Results and Analysis

## *Analysis of Job Adverts*

Using the chrome extension Clay, we collected 158 AI related vacancies in the UK that had been posted on LinkedIn between December 2023 and February 2024. Because of the increasing number of 'work from home' jobs and people accepting longer travel times, the geographical location of universities and companies can often be overlooked, but it is still worth including in analysis. The geographical distribution for the AI related jobs in our dataset is shown in Figure 4.

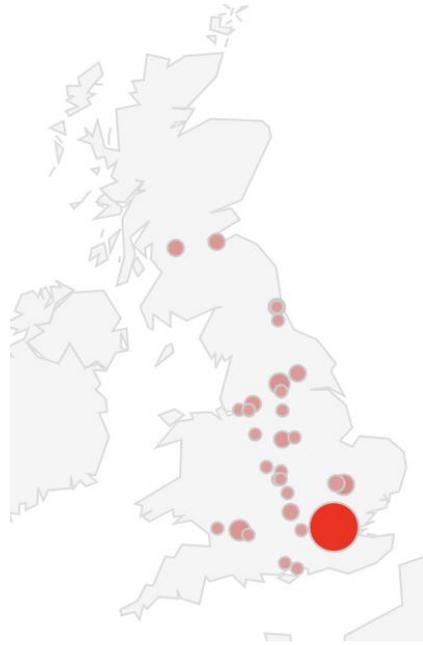

**Figure 4: Geographical locations of most AI job vacancies in the UK.**

The results show the majority of the AI related jobs are offered in London as can be explained by the fact that London is the UK centre for AI start-ups and companies (80% of top 50 AI companies in the UK are located in London (Cheesman, 2024)). Our findings also showed that leading London universities offering AI degrees (such as University College London, Imperial College London and King's College London) have comparatively larger AI and machine learning research groups, leading to the London cluster being significantly larger than others (Hall and Pesenti, 2017). Our map is similar to that created by Open Innovation (with support from Digital Catapult, and Innovate UK (Forth and Billingsley, 2017)) which also shows the distribution of AI research and start-ups. A similar geographical analysis was also seen in studies previously discussed in the Related works section (Department for Digital, Culture, Media and Sport, 2021; Attwood and Williams, 2023).

Another interesting result that we obtained, is that 9.5% of jobs in our study offered a remote working option, so that applicants could potentially be based in any city in the UK. This trend for Artificial Intelligence Engineers is also highlighted in the LinkedIn report mentioned earlier (LinkedIn News UK, 2024).

Our job dataset also revealed information related to the salaries provided. The range is very wide; the lowest being £20,000 per year and the highest around £200,000 per year. It is also noticeable that most of the highest paid jobs are in London while the least paid jobs are in smaller cities such as Plymouth. Though the focus of our study was not on salaries, but rather on identifying the skills gap, the salary range we observed is similar to that observed in the study of Attwood and Williams (2023).

### *Analysis of University Degrees*

Thirty top universities offering a degree in AI were analysed. As mentioned above, 61 out of 166 UK universities offer undergraduate AI courses in 2024 (Figure 5). Further analysis showed that 79 universities offer an MSc degree in AI, amongst which 13 offer an online MSc degree in AI (Figure 6). In total, 46 universities offer both undergraduate and postgraduate degrees in AI. Around 60% of UK universities do offer AI courses, but roughly 40% of these have only been doing so for less than 6 years; thus, we cannot be confident that their teaching standards are comparable to the other courses that they offer. Indeed, the fact that such degrees have been relatively recently launched may deter students from taking these degrees. Lastly, only around 16% of universities in the UK have been offering AI courses for more than 5 years (see Figure 3).

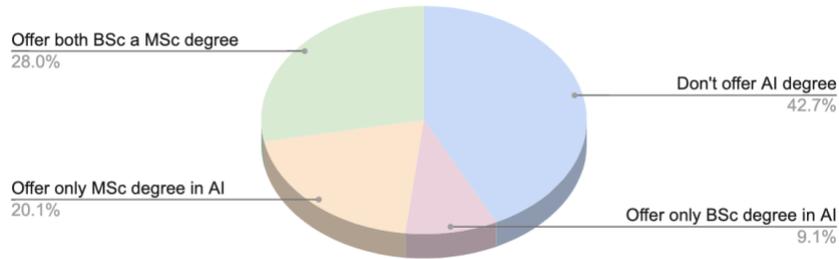

**Figure 5: Proportion of universities that offer AI courses.**

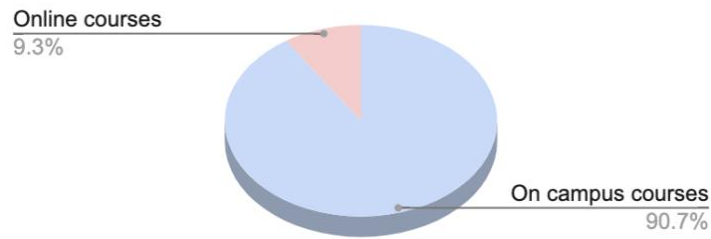

**Figure 6: Proportion of universities programmes on campus and online.**

We observe that the popularity of online programmes is rising, with many universities introducing distant educational degrees that have a flexible active learning format and flipped class approach and making available resources through Virtual Learning Environments (VLE) including virtual laboratories (Kuzminykh et al., 2021; Xiao et al., 2024; Lemeshko et al., 2020). AI degrees are no exception: around 10% of all AI degrees in the UK can be delivered remotely and entirely online. The benefit of online programmes is their flexibility: they improve accessibility for students from different countries, continents and with different backgrounds and experience, and are suitable for working people and others (e.g., carers) with competing commitments.

*Frequency Analysis*

Figure 7 presents the results of our frequency analysis of skills within each category. We can see that the most commonly sought and taught skills are those related to machine learning, programming and software development. Less importance is given to business management and cloud technologies. While research, tools and ethics related skills form the smallest portion of the datasets.

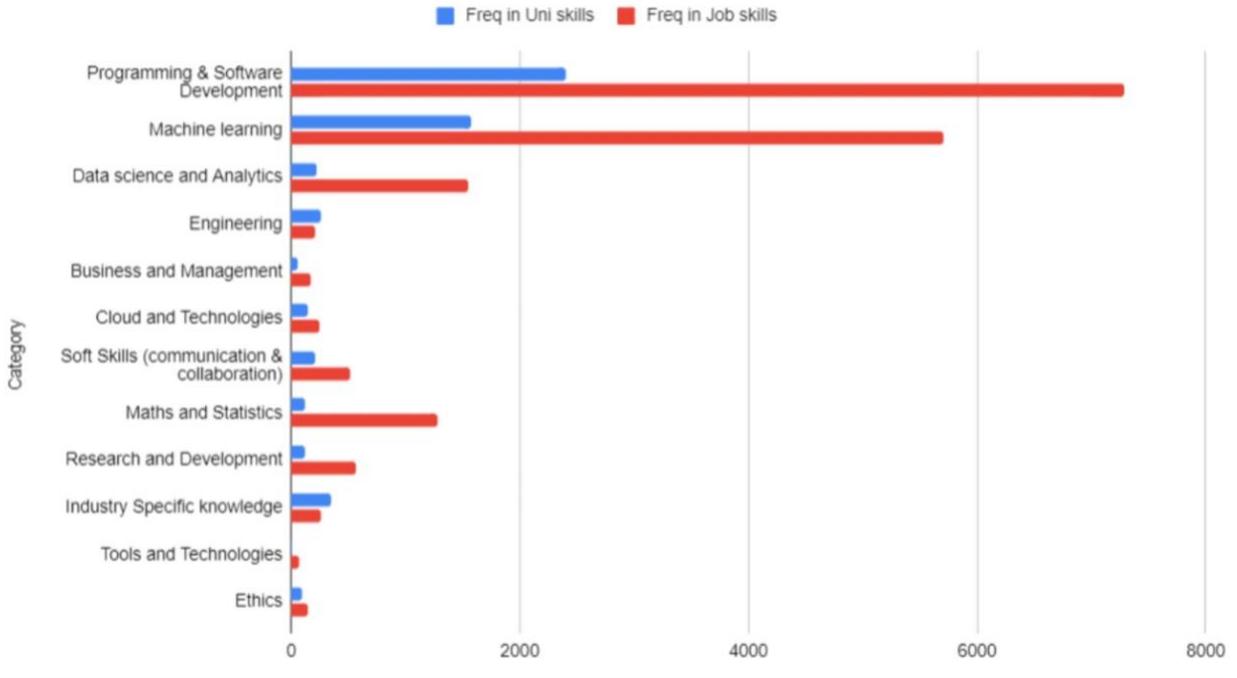

**Figure 7: Frequency analysis of skills under each category.**

## *Mapping Analysis*

To compare the skills offered by universities and required by industry, we use the proportions that each category accounts for in a given dataset. Figure 8 shows pie charts for comparing and visualising each dataset. The abbreviations for categories are taken from Table 1. At first glance, the two pie charts look quite similar, mainly because a large section of both are occupied by either PROG or ML.

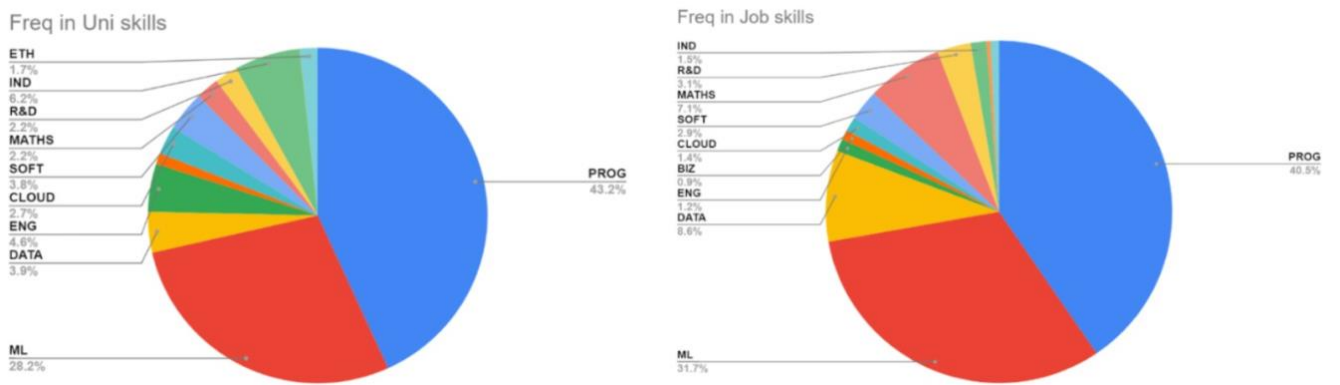

**Figure 8: Frequency analysis of skills under each category.**

A more detailed analysis of these pie charts and skills categories yields more fine-grained inferences, which we summarise by the following key findings:

- *Strong technical skills dominance*. Both in university degrees and AI jobs, the dominant categories are Programming & Software Development and Machine Learning. This highlights the continued and significant demand for strong technical skills in the field of AI.
    - University degrees: The technical skills taught at universities are programming languages like python, Java, C++. On the machine learning side, universities help consolidate concepts such as supervised learning, unsupervised learning and reinforcement learning.
    - AI jobs: Even though the above basic technical skills are required in AI jobs, professional settings usually require more practical skills. Experience with handling and processing large amounts of data is also preferred. This scaling up is a big jump from university, but students equipped with the appropriate general understanding will be able to quickly adapt to meeting the challenges posed by large amounts of data. Strong programming skills are still required, but there is increasing use of additional frameworks such as Keras and Sci-kit learn.
- *Increased demand for data science in jobs*. Data Science and Analytics shows a substantial increase in importance in AI jobs compared to university degrees. This indicates the growing significance of data science skills in practical AI applications.
    - University Degrees: data science at university level typically includes knowledge of languages such as Python, R and SQL.
    - AI Jobs: job postings usually highlight the importance of advanced data science skills such as cloud computing, and big data technologies such as real-time data processing. Familiarity with data processing and visualisation is also preferred. The kinds of projects conducted at university do not currently make use of such advanced technologies.
- *Business and management emphasis in jobs*. Although business and management skills form a relatively low proportion of both datasets, they are more sought for in AI jobs.
    - University Degrees: The soft and managerial skills that are included here can only be gained by joining university electives and academic clubs, it is very unusual to find such aspects being directly taught on an AI course.
    - AI Jobs: However, business analytics is highly valued in the workplace. A role such as an AI consultant is usually filled by someone who not only has a strong technical background but can also make strategic business decisions while working with a client. This gap can only be bridged if universities let students work for external clients for a small duration of time as part of the course.
- *Emphasis on soft skills*. Soft skills are roughly equally emphasised in both skill sets.
    - University Degrees: Soft skills such as communication, collaborating, constructive criticism, etc. are usually gained via group projects at universities.
    - AI jobs: The above-mentioned skills are transferable to team projects that are common in the workplace; hence a large skills difference is not seen here.
- M*aths and statistics significance in jobs*. A high level of quantitative based skills is demanded by job adverts, but somewhat surprisingly, this is not reflected in the extent to which these are being taught at university.

- University Degrees: The maths and statistics taught is very theoretical. Topics such as probability, optimisation techniques, stats and algebra can be found in teaching offerings.
- AI jobs: The practical application of these skills is what is required by the professional industry, and is usually where university students lack experience. E.g. students are able to optimise code in a sophisticated manner but are not able to implement their solution in statistical softwares such as R or tableau, because of a lack of "hands-on" lack of experience.

- *Industry knowledge is marginally more prominent at universities.*
  - University Degree: Surprisingly, while AI jobs still recognize the importance of domain-specific knowledge, this category is slightly more prominent in university degrees. The domain knowledge is usually gained by integrated courses e.g. MSc in Health Data Analytics and Machine Learning (Imperial College London).
  - AI Jobs: Industries usually prefer their employees to be well rounded rather than have specific domain knowledge. Skills specific to a job or industry are usually learnt on the job.
- *Ethics Recognition*. Both sets recognize the importance of Ethics, reflecting the growing awareness of ethical considerations in AI development and deployment.
  - University Degrees: Extra modules are taught at universities to explain to students the ethical considerations that they must consider when dealing with AI and its applications. These modules are usually worth near to no credits but are compulsory to ensure that all students are educated in the ethics of AI.
  - AI Jobs: The importance of ethics is also prominent in the workforce, especially because implementations will be deployed to external clients so that engineers must take into account the considerations of people and ideas from diverse backgrounds, the ethical implications of deployment more broadly, and the need to react to the ever-evolving landscape of AI regulation.

## Discussion

Our research identified 35 skills organized into 12 different skill categories for AI. The categories that we used for classification focus more on technical rather than soft skills. This is a significant difference of our study as compared with a similar study by Verma, Lamsal and Verma (2022) which grouped most of the technical skills under a single "Occupational Attributes" category. Due to the specific aim of our study vis-à-vis performing a mapping with university curricula, our classification omits skills such as communication, interpersonal skills, administrative skills and employee attributes (motivation, time management, confidence)[1].

The AI skills considered in most demand by the UK job market are Programming skills and Machine learning skills, in line with results from the US focussed study by Verma, Lamsal and Verma (2022), and the survey conducted by the UK government (Department for Digital, Culture, Media and Sport, 2021). However, Big data skills and Math & Statistics skills are in less demand in the UK compared to US market: 9% and 7% respectively for AI jobs in the UK and around 33% and 37% respectively for AI jobs in the US.

---

[1] Of course, as indicated later in this article, a separate discussion can be had as to whether such 'soft' skills *should* be emphasised more in technical university courses, given their importance in the jobs sector.

In conclusion, technical skills are taught and also sought in both our datasets; no skills gap can be seen here. However, a surprising finding is that jobs require students to have a strong analytical background, which students graduating from universities may not have (see below). Another important finding is that recruiters want students to have a wide range of general knowledge and strong communication & collaboration skills rather than domain specific knowledge. Overall, we have identified a skills gap, where recruiters do not expect specific domain knowledge, but instead are looking for more well-rounded students with strong analytical skills and who can work collaboratively and communicate well in the workplace.

The result that strong analytical skills are still required by employers is in line with the study of Sarin (2019) which we referred to in the literature review. The observation that the UK job industry places a large focus on analytical knowledge, and that this is often not a strong skill acquired by university graduates, has also been registered in reports published by the McKinsey Global Institute (2019). The studies of Attwood and Williams (2023) and Department for Digital, Culture, Media and Sport (2021) found that there has been a recent increase in the ethical skills that are required by employers. But, in our investigation, ethical skills only cover a small portion of the overall skills that the employers require. This discrepancy can be accounted for by the fact that Attwood and Williams (2023) investigated job skills in cyber security, in contrast to our AI focus. Given the increasing awareness and concerns about the ethical impact of AI, especially Generative AI, we suggest that both universities companies need to place more emphasis on this attribute.

Ethics is also supported in the current initiatives to develop regulatory policies and guardrails for AI development and deployment (UK Parliament, 2024) that industry will expect graduate literacy in AI governance, regulation, and more generally an awareness of potential harms (SHERPA, 2021). In anticipation of this demand, universities are increasingly integrating ethics into computer science and AI education (The Embedded EthiCS, 2024), as well as formulating new degree programmes in AI and philosophy. It remains to be seen, and indeed is of significant import, as to whether these developments, in terms of demand for AI ethics skills and education in AI ethics skills increase in step with each other.

The findings related to the research questions are highlighted in Table 3.

**Table 3: Research questions and findings.**

| Research Question | Findings |
| --- | --- |
| RQ1: What skills are required by UK AI industry recruiters? | The skills required by industry are summarised in Figure 6. The vast majority of skills (72.2%) belongs to occupation attributes related to the technical skills in programming and machine learning algorithms. |
| RQ2: What skills are gained by students who have completed an AI degree in the UK? | The qualifications offered by universities are summarised in Figure 6. The vast majority of skills (71.4%) belongs to occupation attributes related to the technical skills in programming and machine learning algorithms. |
| RQ3: Do university curricula offer a level of knowledge sufficient to match industry requirements for AI related jobs? | There is in general a good balance between skills inculcated by university curricula and those demanded by industry, notably in Programming and Machine learning skills. However, some skills offered by universities require strengthening, including Data Science and Analytics, and Maths and Statistics. Some skills are in less demand by industry, such as Ethics related skills and Industry specific knowledge. Given the recent national and global focus on the ethics of AI, one could argue that both universities and industry should place more emphasis on Ethics skills. |

## Limitations

*Clay*. Although the chrome extension, Clay, is an efficient and easy to use tool that served its purpose for our analysis very well, it has a drawback that only revealed itself after implementation. Clay only works well for LinkedIn, but when it is used to extract data from other job advertising websites, it often leads to erroneous results. Due to this drawback, this report has only considered jobs advertised on LinkedIn, and not other prominent websites such as Glassdoor or Indeed. Hence, if there was any bias, discrimination or randomness present in the adverts posted by LinkedIn, these will heavily impact our current findings. The impact of our research findings would be strengthened if we had used more scrapers on a wider range of courses.

*Classifier*. The classification report shows that the precision level of the classifier is 87%. Although this is quite high, it can still be improved. The classifier implemented in this report is a naive bayes classifier. But another classifier like a random forest could have provided us with more precision. The reason random forests were only considered (but not used) was because of the high complexity levels and the large amount of time that would be taken for implementation. Random forests usually give very accurate results because they use a bunch of decision trees to make predictions. These trees work together to avoid making mistakes, especially when dealing with lots of different factors, making random forests very capable at handling complex data.

## Global perspectives

We identified that the educational programmes in the UK provide quite balanced curricula in AI related subjects to cover industry demands. However, the UK is not exempt from the global shortage of AI professionals and AI skills. Many reports show that the biggest challenge with adoption of AI is a lack of skilled people and difficulty with hiring necessary AI related workforce (Romeo, 2020; Townsend, 2024; Loukides M, 2022). The most recent report from Salesforce (Salesforce, 2024) who asked 600 IT professionals across different industries, including technology, financial services, media and entertainment, manufacturing, retail, healthcare, the public sector in Australia, France, Germany, the United Kingdom, and the United States, and showed that all sectors face a shortage of workers who can effectively apply AI, with the public sector struggling the most.

The way companies prefer to solve the skill gap is by hiring experienced professionals with AI skills; this option is preferable (53%) compared to retaining and retraining current employees (34%). This trend is evident globally, with AI adopters in Canada being 6.2 times more likely to favour replacing employees over retraining, compared to Germany's 1.7 times, possibly due to stringent labour laws (Deloitte, 2019). However, the figures in 2022 for the option whereby companies prefer to retrain existing staff, has doubled compared to figures in 2019 (Deloitte, 2022). This may be due to the companies' investment strategies limiting investment in new staff, improved quality of educational training and courses on AI tools, and issues with hiring AI specialists.

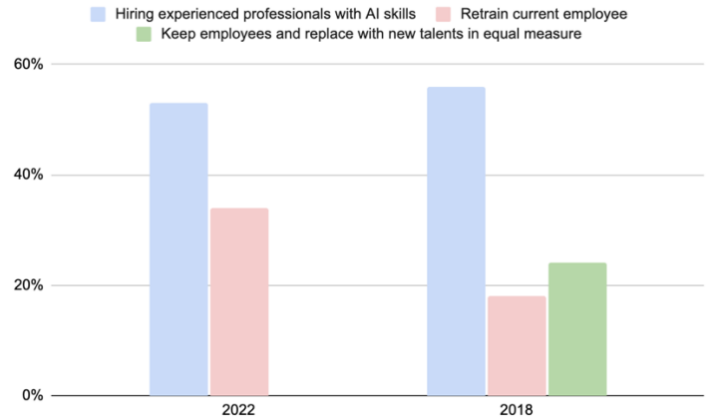

**Figure 9: The strategy of companies to build the AI expertise of employees.**

The strategy to replace workers with new AI skilled professionals is clear, but is it a viable strategy given difficulties recruiting employees with the right expertise? The study of LinkedIn reported a 31.2 million sized AI talent pool in 2022 (Manzyk, 2023). This number is based on the self-reported AI skills on LinkedIn profiles for all levels of expertise. The US has the largest AI talent pool in the world, with over 500,000 people that constitutes 32% of the world pool. The UK, which has over 100,000 people listing AI skills on their LinkedIn profiles, represents the fourth-largest AI talent pool; 6% of the world pool.

At the same time, the number of job adverts seeking recruitment of AI experts indicates the strong demand for AI talent. We performed a search on LinkedIn for AI-based jobs in August 2024 and it yielded more than 236,000 US openings, around 39,000 UK openings and over 700,000 openings worldwide[2]. The size of the AI talent pool is difficult to reconcile with employers' reports of the difficulties filling AI job openings. This contrast can be explained by the fact that while demand for AI experts is high, the supply of people moving into this field with the requisite lefels of qualification and expertise is way below that demanded by companies. It is therefore vital to to ensure not only an appropriate quantity of AI experts, but also that universities prioritise ensuring that their graduates qualify with skills of the requisite quality.

---

[2] Job searches were performed on LinkedIn.com on August 5, 2024, using the Boolean search query: "artificial intelligence" or "ai" or "machine learning" or "deep learning" or "natural language processing" or "computer vision." It's important to note that not all job openings are posted to LinkedIn, and some countries have higher usage of the site than others. We present the numbers as a rough barometer of demand for AI skills.

# Conclusions

The aim of this study was to establish whether there is a skills gap between AI jobs and University AI degrees. To achieve our aim, we scraped data for the 30 top most highly ranked universities in the UK and 158 AI job vacancies in the UK. We then created and used a naive Bayes machine learning model to sort the skills into 12 categories to map the datasets onto each other. We also observed that there is a large proportion of universities that do not currently teach AI modules.

Our findings showed that the most common skills that are taught by universities and also required by industry were those related to Machine Learning, Programming and Software development. There is also a large focus on data science and analytics seen in both datasets. The least common skills are those which are related to research skills, specific tools and technologies and ethics.

Overall, the final mapping showed that the skills offered by universities and required by industry look quite balanced in most of the categories, but the study has revealed a skills gap. Data science and Analytics skills related to data processing, analysis and visualisation showed twice as much demand from industry as compared to what is offered by universities. A similar situation arises with the Mathematics and Statistics category, where the gap is more than 300% according to our results. On the other hand, university curricula offer too much Industry specific knowledge, which is clearly valuable when it comes to understanding real-world scenarios and preparing graduates for careers, but is not valued as an important skill for getting a job. Finally, we suggest that more emphasis should be placed on ethics related skills by both industry and universities, given the wider socio-political and technological emphasis on the importance of understanding the ethical impacts of development and deployment of AI technologies in the coming years.

The results of the study offer valuable insights for use by universities that already teach AI related degrees, as well as those intending to offer AI courses in the near future, as well as insights for keeping curricula content up to date with respect to the requirements of the AI industry.